\documentclass[11pt]{article}

\PassOptionsToPackage{hyperfootnotes=false}{hyperref}
\usepackage[preprint]{acl}

\usepackage{times}
\usepackage{latexsym}
\usepackage[T1]{fontenc}
\usepackage[utf8]{inputenc}
\usepackage{microtype}
\usepackage{inconsolata}
\usepackage{graphicx}
\usepackage{amsmath}
\usepackage{amssymb}
\usepackage{booktabs}
\usepackage{multirow}
\usepackage{algorithm}
\usepackage{algpseudocode}
\usepackage[table]{xcolor}

\title{Query-Guided RoPE Modulation for Lost-in-the-Middle Mitigation in Long-Context LLMs}

\author{
 \textbf{Peiyu Zhang\textsuperscript{1,$\ast$}},
 \textbf{Heng Ping\textsuperscript{1,$\ast$}},
 \textbf{Shinibali Bhattacharyya\textsuperscript{2,$\dagger$}},
 \textbf{Anwesan Pal\textsuperscript{2,$\dagger$}},
 \\
 \textbf{Nikos Kanakaris\textsuperscript{2,$\dagger$}},
 \textbf{Kevin Cheang\textsuperscript{2,$\dagger$}},
 \textbf{Shixuan Li\textsuperscript{1}},
 \textbf{Wei Yang\textsuperscript{1}},
 \textbf{Paul Bogdan\textsuperscript{1}}
\\
 \textsuperscript{1}University of Southern California,
 \textsuperscript{2}Amazon Web Services
}

\begin{document}
\maketitle
\renewcommand{\thefootnote}{$\ast$}
\footnotetext{Equal contribution.}
\renewcommand{\thefootnote}{$\dagger$}
\footnotetext{The work does not relate to the author's position at Amazon.}
\renewcommand{\thefootnote}{\arabic{footnote}}

\begin{abstract}
Long-context language models often neglect information in the middle of the input, a phenomenon known as ``lost-in-the-middle''. Part of this gap may stem from the long-range decay of rotary position embedding (RoPE), which weakens the positional signal between distant tokens. Existing input compression deletes low-importance tokens, and attention reweighting acts only after RoPE has weakened this signal. RoPE-level methods modify positions directly but apply the same remapping to every query. To address these limitations, we propose \textbf{QGPM} (Query-Guided Positional Modulation), a training-free method that reshapes RoPE positions using query relevance from the model's own attention heads. QGPM builds a broad relevance map for lower layers and a sharp one for upper layers. Low-relevance tokens advance by less than one position on fast-rotating planes, pulling relevant evidence toward the query. Slow-rotating planes keep the original layout, with every token in place. Experiments on NoLiMa and RULER with multiple context lengths show that QGPM improves over base models and strong baselines, with the largest gains on retrieval tasks, and generalizes across multiple backbones. Our code is available at \url{https://github.com/iZHpy/QGPM}.
\end{abstract}

\section{Introduction}

Long-context language models now support context windows of hundreds of thousands of tokens~\citep{achiam2023gpt,grattafiori2024llama,yang2024qwen2}. Whether these models actually use the information in such windows, however, remains an open question. Models often attend to the beginning and end of the input while neglecting the middle, a phenomenon known as ``lost-in-the-middle''~\citep{liu2024lost}. Benchmarks designed to probe context utilization~\citep{modarressi2025nolima,hsieh2024ruler} show that accuracy drops sharply as context grows, especially when the relevant evidence shares little surface overlap with the query. This paper therefore aims to improve how models use the information already in the context window, rather than to extend the window further.

Part of this utilization gap may stem from rotary position embedding (RoPE)~\citep{su2024roformer}, the position encoding adopted by most modern LLMs. RoPE represents position as rotations in many two-dimensional planes, each with a distinct frequency. Over a long distance, the fast-rotating planes complete many full turns and no longer carry a consistent distance signal. Evidence far from the query is thus matched through fewer planes than nearby text. This long-range decay depends on the positional distance between tokens. Assigning relevant evidence a position closer to the query can therefore mitigate the decay without changing the input text.

Existing methods that target lost-in-the-middle fall short of intervening on RoPE in a query-aware manner. Input compression methods~\citep{jiang2024longllmlingua,pan2024llmlingua2,li2023compressing} delete low-importance tokens and thus discard graded relevance information. Attention reweighting methods~\citep{zhang2024tell,zhang2024model,guardieiro2025instruction} strengthen attention to selected tokens, but only after RoPE has weakened the positional signal between distant tokens. RoPE-level methods~\citep{zhang2024found,jin2024llm,an2025why} modify position encoding directly, yet apply the same remapping regardless of the query. A separate line of work shows that a few attention heads already attend to query-relevant spans in long inputs~\citep{wu2025retrieval,zhang2025query}. Such attention signals have been used to select KV cache entries~\citep{li2024snapkv}, filter input tokens~\citep{shi2026discovering}, or re-rank documents~\citep{zhang2025query}, but always by deleting or reordering content. In short, no existing method uses the model's own query relevance to reshape RoPE positions while keeping every token in place.

We propose \textbf{QGPM} (Query-Guided Positional Modulation), a training-free method that fills this gap. QGPM reads the attention from the query to each context sentence in a few such heads. The resulting relevance ranking is shaped into a broad map and a sharp map. Using the two maps, QGPM lets low-relevance tokens advance by less than one position on the fast-rotating planes, so relevant sentences move closer to the query. In contrast, the slow-rotating planes keep the original layout of the document. The broad map is applied to the lower layers and the sharp map to the upper layers. QGPM deletes or reorders no token and updates no model weight.

\begin{figure*}[t]
  \centering
  \includegraphics[width=\textwidth]{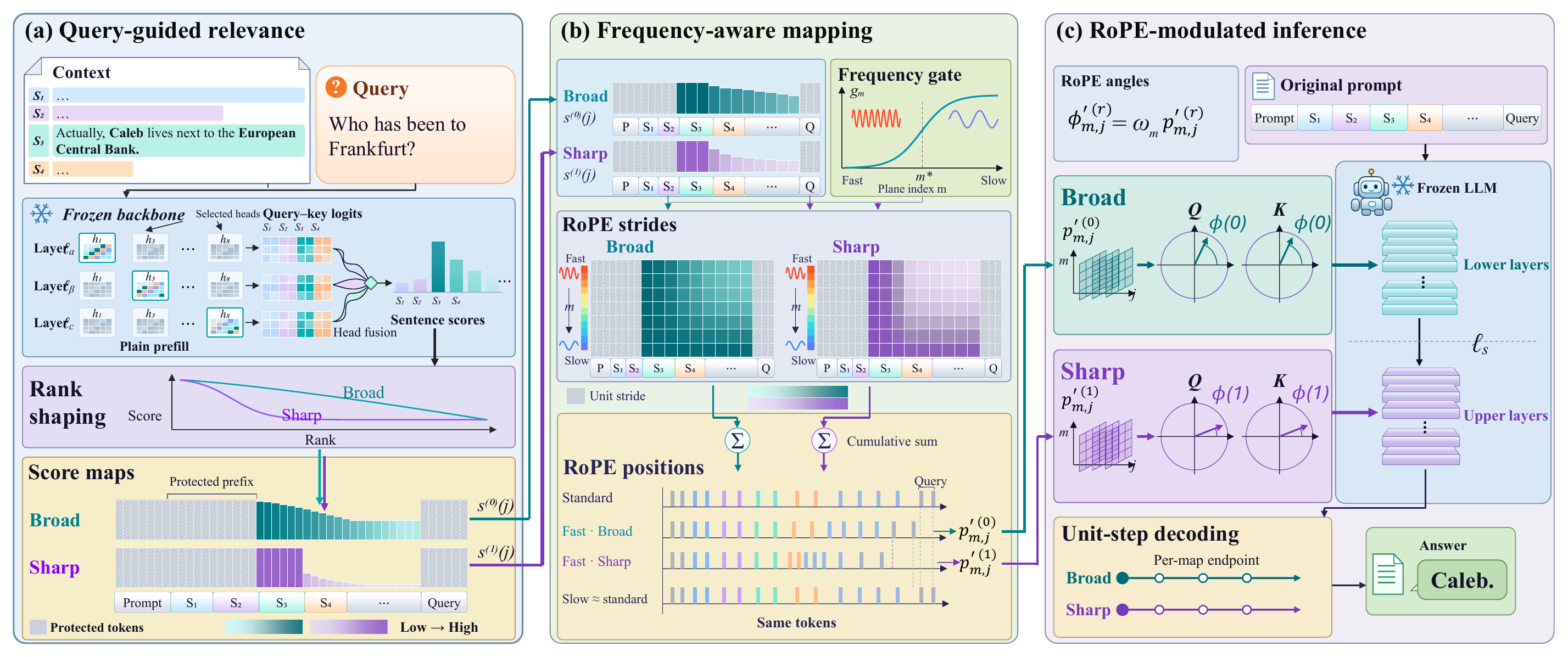}
  \caption{Overview of QGPM. (a) A plain prefill reads the attention from the query to each context sentence in a few selected heads. The fused sentence ranking is shaped into a broad relevance map and a sharp score map. Context before the first relevant sentence is protected and keeps a score of 1. (b) Each map sets a positional stride for every token on every frequency plane. Low-relevance tokens advance by less than one position on fast planes. Slow planes, relevant sentences, and the query keep unit stride. Cumulative sums give the remapped positions. (c) In the answering pass, the lower layers use the broad map and the upper layers use the sharp map. Decoding continues in unit steps from the last position of each map.}
  \label{fig:pipeline}
\end{figure*}

Our main contributions are as follows:

\begin{itemize}
    \item We propose QGPM, which converts query relevance read from the model's own attention heads into token-level and frequency-specific RoPE positions. The same context thus receives different positions under different queries, which makes the evidence needed by each query easier for the model to retrieve.
    \item We design a modulation scheme that is aware of both frequency and depth. Low-relevance text is compressed only on fast-rotating planes, and separate relevance maps guide the lower and upper layers.
    \item Comprehensive experiments on multiple LLM backbones and long-context benchmarks show that QGPM achieves superior overall performance over strong baselines across a range of context lengths.
\end{itemize}

\section{Related Work}
\label{sec:related-work}

\subsection{Mitigating Lost-in-the-Middle}

Inference-time methods for lost-in-the-middle~\citep{liu2024lost} mostly edit the input or the attention weights without touching position encoding. Input compression methods such as Selective Context~\citep{li2023compressing}, LongLLMLingua~\citep{jiang2024longllmlingua}, and LLMLingua-2~\citep{pan2024llmlingua2} delete tokens judged unimportant. TAG~\citep{pal2025tagging} instead inserts tags that mark salient content. Attention reweighting methods such as PASTA~\citep{zhang2024tell}, AutoPASTA~\citep{zhang2024model}, and InstABoost~\citep{guardieiro2025instruction} amplify attention to selected tokens, and \citet{hsieh2024found} calibrate the positional bias in attention. Deletion loses information, and tagging leaves the long-range decay untouched. Attention-side methods can only rescale a positional signal that RoPE has already weakened.

\subsection{RoPE and Position Modification}

Most RoPE modifications extend the context window by rescaling position indices, as in Position Interpolation~\citep{chen2023extending}, YaRN~\citep{peng2024yarn}, and LongRoPE~\citep{ding2024longrope}. Fewer methods improve utilization within the window. Ms-PoE~\citep{zhang2024found} rescales positions with head-specific ratios, and Self-Extend~\citep{jin2024llm} maps distant positions back into the trained range by grouping. STRING~\citep{an2025why} reuses well-trained positions for long distances, and Position Engineering~\citep{he2024position} inserts placeholder positions tuned for each task. All of these remappings are fixed before the query is seen, whereas QGPM derives the remapping from the query.

\subsection{Attention as a Relevance Signal}

Attention itself provides a reliable relevance signal in long inputs. A sparse set of retrieval heads copies information from long contexts~\citep{wu2025retrieval}, and QRHead~\citep{zhang2025query} uses query-focused heads to re-rank documents. Attention sorting~\citep{peysakhovich2023attention} reorders documents by attention, while SnapKV~\citep{li2024snapkv} and GemFilter~\citep{shi2026discovering} use attention to select KV cache entries and input tokens, respectively. All of these methods use the signal to select or reorder content. QGPM instead uses a similar query-to-context signal to modulate RoPE positions, so every token stays in the input.

\section{Preliminaries}
\label{sec:preliminary}

\subsection{Rotary Position Embedding}

RoPE~\citep{su2024roformer} splits each query and key vector of dimension $d$ into $d/2$ two-dimensional planes. Plane $m$ rotates with angular frequency
\begin{equation}
\omega_m = \theta^{-2m/d}, \quad m = 0, 1, \ldots, d/2 - 1,
\label{eq:rope-freq}
\end{equation}
where $\theta$ is a base constant, typically between $10^4$ and $10^8$. Plane $m = 0$ rotates fastest. A token at position $p$ is rotated by $R(p) = \mathrm{diag}\big(R_2(\omega_0 p), \ldots, R_2(\omega_{d/2-1} p)\big)$, where $R_2(\phi)$ is a planar rotation by angle $\phi$. For a query vector $q_i$ at position $p_i$ and a key vector $k_j$ at position $p_j$, the identity $R(p_i)^\top R(p_j) = R(p_j - p_i)$ gives
\begin{equation}
\begin{aligned}
\mathrm{logit}(i, j) &= q_i^\top R(-\Delta p)\, k_j \\
&= \sum_{m=0}^{d/2-1} \big[\, a_m \cos(\omega_m \Delta p) \\
&\qquad\quad + b_m \sin(\omega_m \Delta p) \,\big],
\end{aligned}
\label{eq:rope-expansion}
\end{equation}
where $\Delta p = p_i - p_j$, $a_m = q_{i,2m} k_{j,2m} + q_{i,2m+1} k_{j,2m+1}$, and $b_m = q_{i,2m} k_{j,2m+1} - q_{i,2m+1} k_{j,2m}$. The logit thus depends on position only through $\Delta p$, and each plane adds one oscillating term.

\subsection{Frequency Planes and Long-Range Decay}

Plane $m$ completes one full turn every $\lambda_m = 2\pi/\omega_m$ positions, and $\lambda_m$ grows exponentially with $m$. At a distance $\Delta p$, every plane with $\lambda_m < \Delta p$ has turned at least once. The terms of these planes in Equation~\ref{eq:rope-expansion} no longer change consistently with distance and tend to average out across nearby keys. Only planes with $\lambda_m > \Delta p$ keep a coherent distance signal, namely the planes above the boundary index
\begin{equation}
m(\Delta p) = \frac{d \ln(\Delta p / 2\pi)}{2 \ln \theta}.
\label{eq:wrap-index}
\end{equation}
Since $m(\Delta p)$ grows with $\Delta p$, distant evidence is matched through fewer planes than nearby text. For example, with $\theta = 10^7$ and $d = 128$, about 53 of 64 planes stay coherent at 100 tokens, but only about 30 at 32K tokens. We refer to this effect as the long-range decay of RoPE.

The same view suggests how to counter the decay. Shrinking the distance between the query and relevant evidence from $\Delta p$ to $\Delta p'$ restores coherence on every plane with $\Delta p' < \lambda_m < \Delta p$. Slow planes with wavelengths beyond the context length stay coherent at any distance and keep the coarse layout of the document. These planes need no change. Fast planes also encode local word order, so changes on fast planes should be limited to text that is irrelevant to the query.

\section{Method: QGPM}
\label{sec:method}

QGPM runs the backbone twice on each input (Figure~\ref{fig:pipeline}). A plain prefill first measures how strongly the query attends to each context sentence (Section~\ref{sec:method-scoring}). The resulting ranking is shaped into a broad and a sharp relevance map (Section~\ref{sec:method-maps}). In the answering pass, each map sets plane-specific strides (Section~\ref{sec:method-stride}), and the accumulated strides replace the original positions in RoPE (Section~\ref{sec:method-remapping}).

\subsection{Relevance from Attention Heads}
\label{sec:method-scoring}

QGPM reads relevance from a small set $\mathcal{H}$ of attention heads. The heads are selected once per backbone by how well each head ranks annotated evidence on a held-out set (Appendix~\ref{app:heads}). Let $\mathcal{Q}$ denote the query rows, namely the question tokens and the last few prompt tokens (Appendix~\ref{app:implementation}). For head $h$ and context token $j$, the relevance signal is the largest RoPE logit from any query row:
\begin{equation}
\ell_h(j) = \max_{r \in \mathcal{Q}} \frac{\big(R(p_r)\, q_{h,r}\big)^\top \big(R(p_j)\, k_{h,j}\big)}{\sqrt{d}}.
\label{eq:head-signal}
\end{equation}
QGPM splits the context into sentences $S_1, \ldots, S_n$ and scores each sentence by the strongest token in the sentence, $\ell_h(S_i) = \max_{j \in S_i} \ell_h(j)$. Each head is standardized over sentences, and the heads are averaged:
\begin{equation}
z_i = \frac{1}{|\mathcal{H}|} \sum_{h \in \mathcal{H}} \frac{\ell_h(S_i) - \mu_h}{\sigma_h},
\label{eq:fusion}
\end{equation}
where $\mu_h$ and $\sigma_h$ are the mean and standard deviation of $\ell_h(S_i)$ over sentences.

\subsection{Broad and Sharp Relevance Maps}
\label{sec:method-maps}

The scale of $z_i$ varies across inputs, so QGPM keeps only the order of the sentences. Let $\rho(i)$ be the rank of $S_i$ by $z_i$, with $\rho(i) = 0$ for the most relevant sentence. The score of $S_i$ is
\begin{equation}
\begin{aligned}
f_i &= \sigma\big((K - 0.5 - \rho(i)) / T\big), \\
s_i &= s_{\mathrm{low}} + (1 - s_{\mathrm{low}})\, \tilde f_i,
\end{aligned}
\label{eq:rank-scoring}
\end{equation}
where $\sigma(\cdot)$ is the sigmoid function and $\tilde f_i$ rescales $f_i$ to $[0, 1]$ over sentences. The top sentence thus receives a score of 1, and the bottom sentence receives $s_{\mathrm{low}}$. The center $K$ sets how many sentences stay near 1, and the temperature $T$ sets how quickly the scores fall off. Setting $T = 0$ gives a hard top-$K$ cut.

QGPM shapes the same ranking twice. The broad map uses a larger $K$ and a softer $T$, so several candidate sentences stay relevant. The sharp map uses a smaller $K$ and a harder $T$, so only the top sentences stay relevant. Each token inherits the score of the enclosing sentence. Tokens outside the context, such as the instruction and the question, always receive a score of 1.

We call a sentence relevant if the sentence score is at least $(1 + s_{\mathrm{low}})/2$. In each map, all context tokens before the first relevant sentence also receive a score of 1 and form a protected prefix. Under causal attention, the representation of the evidence is built from the text before the evidence. Keeping this prefix intact preserves the encoding of the evidence. Compression is then limited to the text between the evidence and the query, which is the span that sets the distance between the two.

\begin{algorithm}[t]
\caption{QGPM inference}
\label{alg:qgpm}
\begin{algorithmic}[1]
\Require context sentences $S_1, \ldots, S_n$, heads $\mathcal{H}$, query rows $\mathcal{Q}$
\Require hyperparameters $(K_b, T_b)$, $(K_s, T_s)$, $s_{\mathrm{low}}$, $L_a$, $\tau_m$, $\epsilon$
\Statex \textit{Pass 1: relevance scoring}
\State Run a plain prefill up to the deepest layer in $\mathcal{H}$
\State Compute $\ell_h(j)$ and $z_i$ by Eqs.~\ref{eq:head-signal} and~\ref{eq:fusion}
\State Rank the sentences by $z_i$ to obtain $\rho(i)$
\For{each map $g \in \{\text{broad}, \text{sharp}\}$}
    \State Score sentences by Eq.~\ref{eq:rank-scoring} with $(K_g, T_g)$
    \State Set scores of non-context tokens and the protected prefix to 1
    \State Compute strides by Eq.~\ref{eq:stride} with $m^* = m(L_a)$
    \State Accumulate positions $p'_{m,j}$ by Eq.~\ref{eq:cumsum}
\EndFor
\Statex \textit{Pass 2: answering}
\State Prefill with the broad map below layer $\ell_s$ and the sharp map from $\ell_s$ on
\State Decode in unit steps from $p'_{m,N-1}$ of each map
\end{algorithmic}
\end{algorithm}

\subsection{Frequency-Aware Strides}
\label{sec:method-stride}

Following Section~\ref{sec:preliminary}, QGPM changes only the fast planes. Given an anchor length $L_a$ proportional to the context length, the boundary plane is $m^* = m(L_a)$ from Equation~\ref{eq:wrap-index}. Planes with $m < m^*$ complete at least one turn within $L_a$ tokens and count as fast. The remaining planes count as slow. A sigmoid gate of width $\tau_m$ separates the two groups. For a map with token scores $s(j)$, the stride $\delta_m(j)$ of token $j$ on plane $m$ is
\begin{equation}
\begin{aligned}
g_m &= \sigma\big((m - m^*) / \tau_m\big), \\
\delta_m(j) &= \max\big(s(j) + (1 - s(j))\, g_m,\ \epsilon\big),
\end{aligned}
\label{eq:stride}
\end{equation}
where $\epsilon$ is a stride floor. On slow planes $g_m \approx 1$, so every token keeps unit stride. On fast planes $g_m \approx 0$, so the stride falls to about $\max(s(j), \epsilon)$. Relevant sentences, whose scores are close to 1, thus keep near-unit stride on every plane, and only low-relevance tokens on fast planes are compressed.

The floor $\epsilon$ keeps fast-plane increments close to the range seen in pretraining. With a small floor, about $1/\epsilon$ consecutive low-relevance tokens share one position on each fast plane, so the local order of these tokens is largely lost. We hypothesize that this blur de-emphasizes distractor text without deleting the text, while the slow planes still place every token correctly within the document.

\subsection{Remapped Positions and Layer Schedule}
\label{sec:method-remapping}

Remapped positions accumulate the strides:
\begin{equation}
p'_{m,j} = \sum_{k=1}^{j} \delta_m(k).
\label{eq:cumsum}
\end{equation}
Every stride is at least $\epsilon > 0$, so positions stay strictly increasing. Token $j$ is rotated by $\omega_m p'_{m,j}$ on plane $m$, and the logit follows Equation~\ref{eq:rope-expansion} with $\Delta p$ replaced by the plane-specific offset $\Delta p'_{m,ij} = p'_{m,i} - p'_{m,j}$. Because positions accumulate, unit stride does not mean an unchanged position. On fast planes, every token after a compressed span moves toward the span, which brings relevant sentences closer to the query. Distances within the query are preserved exactly, and distances within relevant sentences are preserved approximately.

QGPM computes one set of positions for each map. Layers below $\ell_s = \lfloor n_{\mathrm{layer}}/2 \rfloor$ use the broad map, and the remaining layers use the sharp map. Each layer stores the KV cache in the coordinates of the corresponding map. We hypothesize a division of labor across depth. Lower layers build token representations from local context, so text blurred there cannot be recovered later. The broad map therefore keeps several candidate sentences intact in these layers. The selected heads lie mostly in the middle and upper layers, where the sharp map leaves fewer competing sentences near the query.

During decoding, the $k$-th generated token takes position $p'_{m,N-1} + k$ on each plane of each map, where $N$ is the prompt length. Decoding thus continues in unit steps from the last prompt position. Compared with standard inference, QGPM adds one plain prefill that stops at the deepest selected layer. The per-token cost of decoding stays equal to that of standard RoPE (Appendix~\ref{app:cost}). Algorithm~\ref{alg:qgpm} summarizes the full procedure, and Appendix~\ref{app:hyperparams} describes all hyperparameters.

\section{Experiments}
\label{sec:experiments}

\begin{table*}[t]
\centering
\small
\setlength{\tabcolsep}{3pt}
\begin{tabular*}{\textwidth}{@{\extracolsep{\fill}} l cccccc @{\hspace{12pt}} cccccc @{}}
\toprule
 & \multicolumn{6}{c}{Qwen2.5-7B} & \multicolumn{6}{c}{Qwen3-4B} \\
\cmidrule(lr){2-7} \cmidrule(l){8-13}
Method & NoLiMa & QA-1 & QA-2 & MK & MV & Avg. & NoLiMa & QA-1 & QA-2 & MK & MV & Avg. \\
\midrule
Vanilla & 28.5 & 79.6 & 57.0 & 94.3 & 83.5 & 68.6 & 16.1 & 77.5 & 56.6 & 99.6 & 96.0 & 69.2 \\
LLMLingua-2 & 19.7 & 51.0 & 40.0 & \phantom{0}2.2 & 65.3 & 35.6 & 10.4 & 49.1 & 36.6 & \phantom{0}2.0 & 75.9 & 34.8 \\
Ms-PoE & 32.9 & 77.2 & 52.0 & 36.8 & 68.5 & 53.5 & 16.6 & 75.4 & 55.3 & 88.2 & 84.7 & 64.0 \\
Self-Extend & 29.5 & 79.3 & 57.5 & 78.8 & 82.6 & 65.5 & 18.1 & 76.5 & 57.5 & \textbf{99.8} & 92.2 & 68.8 \\
STRING & 33.1 & 79.4 & 57.8 & 92.3 & 76.6 & 67.8 & 22.9 & 78.8 & \textbf{60.0} & 99.3 & 93.0 & 70.8 \\
\midrule
QGPM & \textbf{46.1} & \textbf{80.4} & \textbf{58.0} & \textbf{99.4} & \textbf{93.4} & \textbf{75.4} & \textbf{25.1} & \textbf{80.1} & 56.2 & 99.6 & \textbf{96.8} & \textbf{71.6} \\
\midrule[\heavyrulewidth]
\addlinespace[2pt]
 & \multicolumn{6}{c}{Ministral-8B$^{\dagger}$} & \multicolumn{6}{c}{Granite-3.3-8B} \\
\cmidrule(lr){2-7} \cmidrule(l){8-13}
Method & NoLiMa & QA-1 & QA-2 & MK & MV & Avg. & NoLiMa & QA-1 & QA-2 & MK & MV & Avg. \\
\midrule
Vanilla & 11.5 & 73.5 & 59.0 & 89.4 & 91.4 & 65.0 & 18.6 & 78.6 & 62.7 & 66.0 & 71.1 & 59.4 \\
LLMLingua-2 & \phantom{0}6.2 & 48.2 & 40.8 & \phantom{0}3.6 & 78.0 & 35.3 & 14.1 & 50.1 & 43.1 & \phantom{0}0.2 & 48.4 & 31.2 \\
Ms-PoE & \phantom{0}9.0 & 68.5 & 53.0 & 32.4 & 81.2 & 48.8 & 18.5 & 74.1 & 60.2 & 61.5 & 14.1 & 45.7 \\
Self-Extend & 11.1 & 72.8 & 58.8 & 94.6 & 92.8 & 66.0 & 16.6 & 77.3 & 61.9 & 56.8 & 71.8 & 56.9 \\
STRING & 17.1 & 73.3 & \textbf{59.1} & 94.7 & 90.8 & 67.0 & 25.9 & \textbf{78.8} & 62.9 & 73.5 & 75.8 & 63.4 \\
\midrule
QGPM & \textbf{24.1} & \textbf{73.6} & 59.0 & \textbf{95.2} & \textbf{93.4} & \textbf{69.1} & \textbf{35.6} & \textbf{78.8} & \textbf{63.1} & \textbf{85.1} & \textbf{78.9} & \textbf{68.3} \\
\bottomrule
\end{tabular*}
\caption{Accuracy (\%) averaged over context lengths, from 4K to 32K for NoLiMa and from 4K to 64K for RULER. MK and MV denote NIAH-MK and NIAH-MV, and Avg.\ is the mean over the five tasks. Bold marks the best result for each backbone. $^{\dagger}$Includes 64K, which exceeds the configured window of Ministral-8B. Appendix~\ref{app:full-results} gives the results at each length.}
\label{tab:main}
\end{table*}

\begin{figure*}[t]
  \centering
  \includegraphics[width=\textwidth]{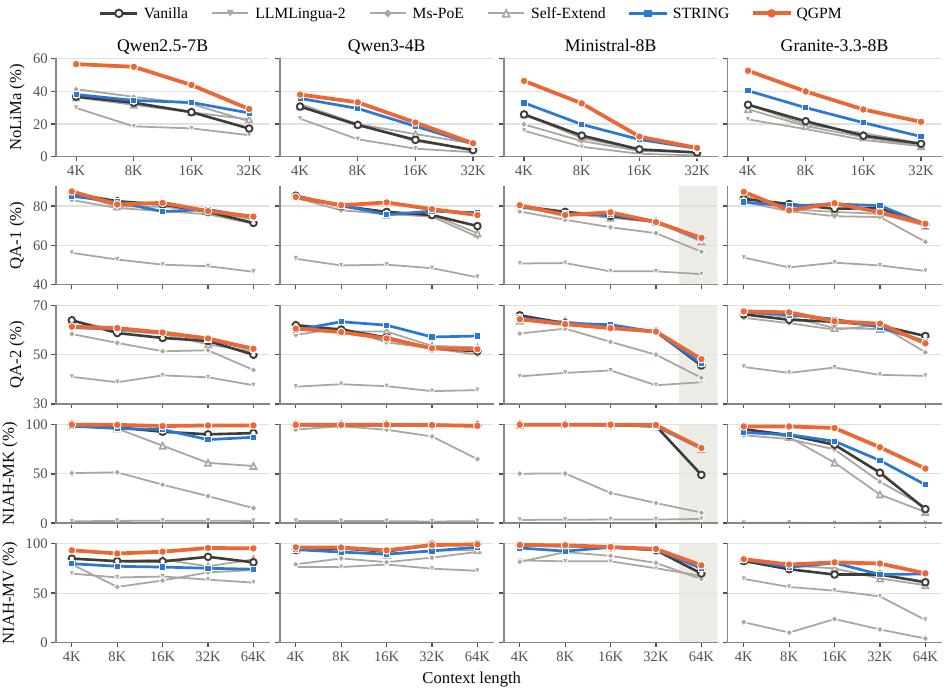}
  \caption{Accuracy (\%) of all five tasks at each context length. Gray lines show LLMLingua-2, Ms-PoE, and Self-Extend. The shaded region marks 64K for Ministral-8B, which exceeds the configured window of this backbone.}
  \label{fig:length}
\end{figure*}

We organize the experiments around three research questions.
\begin{itemize}
    \setlength{\itemsep}{0pt}
    \setlength{\parskip}{0pt}
    \item \textbf{RQ1:} Does QGPM outperform existing inference-time methods across backbones, tasks, and context lengths?
    \item \textbf{RQ2:} Is modulating RoPE positions more effective than selecting or reordering content by relevance?
    \item \textbf{RQ3:} How close does the relevance read from attention heads come to gold evidence?
\end{itemize}
RQ2 and RQ3 examine the two design choices of QGPM. RQ2 concerns how the relevance is used, and RQ3 concerns where the relevance comes from.

\subsection{Experimental Setup}
\label{sec:exp-setup}

\paragraph{Models.}
We evaluate four instruction-tuned backbones from three model families: Qwen2.5-7B-Instruct-1M~\citep{yang2025qwen251m}, Qwen3-4B-Instruct-2507~\citep{yang2025qwen3}, Ministral-8B-Instruct-2410~\citep{mistral2024ministraux}, and Granite-3.3-8B-Instruct~\citep{ibm2025granite33}. The RoPE bases of the four backbones range from $5 \times 10^6$ to $10^8$. Except for Ministral-8B at 64K,\footnote{For Ministral-8B, 64K exceeds the 32,768-token limit in its HuggingFace configuration (\texttt{max\_\allowbreak position\_\allowbreak embeddings} and \texttt{sliding\_\allowbreak window}). Results at this length therefore reflect behavior beyond the configured window and are reported for completeness.} all inputs fit within the native context window of each backbone, and we apply no length extrapolation.

\paragraph{Benchmarks and metrics.}
NoLiMa~\citep{modarressi2025nolima} tests retrieval when the evidence shares little lexical overlap with the question. We evaluate NoLiMa at 4K, 8K, 16K, and 32K tokens. From RULER~\citep{hsieh2024ruler}, we use two question answering tasks, single-hop QA-1 and multi-hop QA-2, and two retrieval tasks, NIAH-MK and NIAH-MV. We evaluate the RULER tasks at 4K, 8K, 16K, 32K, and 64K tokens. NIAH-MK hides one target key among many similar keys, and NIAH-MV asks for all values of one key. We report accuracy (\%) on every task. Correctness is judged by whether the response contains the required gold answers (Appendix~\ref{app:implementation}).

\paragraph{Baselines.}
We compare QGPM with the unmodified backbone, denoted Vanilla, and with four inference-time methods. LLMLingua-2~\citep{pan2024llmlingua2} represents input compression. Ms-PoE~\citep{zhang2024found}, Self-Extend~\citep{jin2024llm}, and STRING~\citep{an2025why} modify RoPE positions without using the query. All baselines use the default settings recommended in the original papers (Appendix~\ref{app:implementation}).

\paragraph{Implementation details.}
All methods use greedy decoding and run with PyTorch and HuggingFace Transformers on NVIDIA A100 GPUs. Appendix~\ref{app:heads} describes head selection, and Appendix~\ref{app:hyperparams} describes the hyperparameters for each backbone and dataset.

\subsection{Main Results (RQ1)}
\label{sec:exp-main}

\paragraph{Overall performance.}
Table~\ref{tab:main} reports the accuracy averaged over context lengths. QGPM achieves the best five-task average on all four backbones and improves on Vanilla by 2.4 to 8.9 points. QGPM is also the only method whose five-task average never falls below Vanilla. Among the baselines, STRING exceeds Vanilla on three backbones and Self-Extend on one. Ms-PoE and LLMLingua-2 exceed Vanilla on none.

\begin{table}[t]
\centering
\small
\setlength{\tabcolsep}{3pt}
\begin{tabular*}{\columnwidth}{@{\extracolsep{\fill}} l cccc @{}}
\toprule
QGPM vs. & NoLiMa & NIAH & QA & All \\
\midrule
Vanilla & 16/0/0 & 31/6/3 & 24/2/14 & 71/8/17 \\
LLMLingua-2 & 16/0/0 & 40/0/0 & 40/0/0 & 96/0/0 \\
Ms-PoE & 16/0/0 & 40/0/0 & 38/1/1 & 94/1/1 \\
Self-Extend & 16/0/0 & 34/3/3 & 26/2/12 & 76/5/15 \\
STRING & 16/0/0 & 38/2/0 & 22/2/16 & 76/4/16 \\
\bottomrule
\end{tabular*}
\caption{Wins, ties, and losses (W/T/L) of QGPM against each method over settings of backbone, task, and context length. The four columns cover 16, 40, 40, and 96 settings.}
\label{tab:wtl}
\end{table}

\paragraph{Retrieval tasks.}
The largest gains appear on retrieval. On NoLiMa, QGPM is the best method at every context length on every backbone. The average gain over Vanilla ranges from 9.0 to 17.6 points. On NIAH, the gains are large where the backbone degrades. QGPM raises NIAH-MK on Granite from 66.0 to 85.1 and NIAH-MV on Qwen2.5 from 83.5 to 93.4. Qwen3 already reaches at least 96.0 on both NIAH tasks, and QGPM stays within one point of Vanilla on this backbone.

\paragraph{Question answering.}
On RULER QA, QGPM performs on par with Vanilla. The average difference ranges from $-0.4$ to $+2.6$ points, and the only drop occurs on QA-2 for Qwen3, where STRING performs best. Section~\ref{sec:exp-oracle} shows that even annotated evidence helps little on QA-1.

\paragraph{Robustness across settings.}
Table~\ref{tab:wtl} compares QGPM with each method in every setting of backbone, task, and context length. QGPM wins in at least 71 of the 96 settings against every method. QGPM never loses on NoLiMa and never loses to STRING on NIAH. Most losses occur on QA. Against Vanilla, QGPM loses in 17 settings by at most 3.2 points, and 14 of these settings are QA.

\paragraph{Baseline behavior.}
LLMLingua-2 falls below Vanilla on every task and nearly fails on NIAH-MK, with averages from 0.2 to 3.6. Deleting tokens likely breaks the exact keys that NIAH-MK must match. The query-agnostic RoPE methods are unstable on retrieval. Ms-PoE averages 14.1 on NIAH-MV for Granite, against 71.1 for Vanilla. Self-Extend drops to 58.0 on NIAH-MK at 64K for Qwen2.5, where Vanilla keeps 91.4. A fixed remapping treats evidence and distractors alike, which may explain these drops. STRING is the strongest baseline, yet the five-task average of STRING still falls below Vanilla on Qwen2.5.

\paragraph{Effect of context length.}
Figure~\ref{fig:length} shows the accuracy of every task at each context length. On NoLiMa, QGPM stays above every baseline at every length. The margin over the strongest baseline peaks at 4K or 8K. At 32K, the margin shrinks to between 0.3 and 2.4 points on Qwen2.5, Qwen3, and Ministral. Granite still keeps a margin of 8.9 points. Section~\ref{sec:exp-oracle} examines the smaller margin at 32K. On NIAH-MK, the gain over Vanilla grows with length on backbones that degrade. On Granite, the gain rises from 2.8 points at 4K to 41.2 points at 64K. On NIAH-MV, QGPM improves on Vanilla at every length on Qwen2.5, by 7.8 to 14.2 points. On QA, QGPM tracks Vanilla closely, and small gains appear mainly at 64K, where QGPM exceeds Vanilla in seven of eight settings.

\subsection{Modulation versus Input Editing (RQ2)}
\label{sec:exp-editing}

\begin{table}[t]
\centering
\small
\setlength{\tabcolsep}{3pt}
\begin{tabular*}{\columnwidth}{@{\extracolsep{\fill}} l ccc @{\hspace{8pt}} cc @{}}
\toprule
 & \multicolumn{3}{c}{Qwen2.5-7B} & \multicolumn{2}{c}{NoLiMa} \\
\cmidrule(lr){2-4} \cmidrule(l){5-6}
Method & NoLiMa & QA-1 & MV & Ministral & Granite \\
\midrule
Vanilla & 28.5 & 79.6 & 83.5 & 11.5 & 18.6 \\
STRING & 33.1 & 79.4 & 76.6 & 17.1 & 25.9 \\
\midrule
Top-K & 41.1 & \textbf{80.6} & 91.1 & n/a & n/a \\
Reorder & 39.9 & 80.1 & 92.2 & n/a & n/a \\
QGPM & \textbf{46.1} & 80.4 & \textbf{93.4} & \textbf{24.1} & \textbf{35.6} \\
\midrule
\textcolor{gray}{Oracle} & \textcolor{gray}{47.3} & \textcolor{gray}{81.6} & \textcolor{gray}{93.9} & \textcolor{gray}{n/a} & \textcolor{gray}{n/a} \\
\bottomrule
\end{tabular*}
\caption{Average accuracy (\%) of QGPM, input editing, and Oracle. MV denotes NIAH-MV, and n/a marks settings that are not run. Top-K and Reorder rely on Qwen3-Reranker-4B, whose tokenizer is not fully compatible with those of Ministral-8B and Granite-3.3-8B. Oracle (gray) uses evidence positions marked by hand, which are costly to obtain for every backbone. Bold marks the best result other than Oracle.}
\label{tab:ablation}
\end{table}

\begin{figure*}[t]
  \centering
  \includegraphics[width=\textwidth]{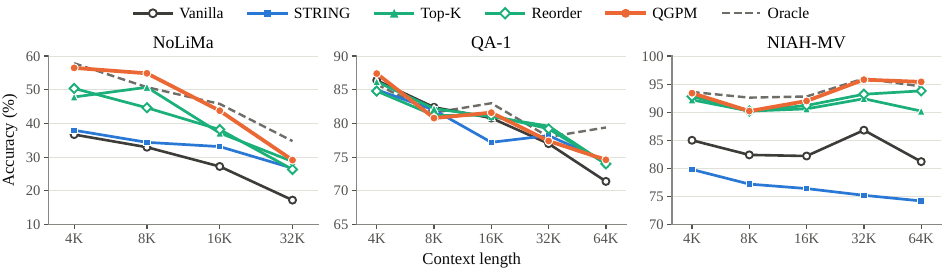}
  \caption{Accuracy (\%) of QGPM, input editing, and Oracle at each context length on Qwen2.5-7B. Top-K and Reorder share one color as the two input editing variants. Oracle (dashed) uses annotated evidence and serves as a reference.}
  \label{fig:edit-length}
\end{figure*}

We compare QGPM with two variants that edit the input according to relevance. Both variants split the context into chunks and score each chunk against the question with Qwen3-Reranker-4B~\citep{zhang2025qwen3embedding}. Top-K keeps the five highest-scoring chunks and drops the rest. Reorder keeps all chunks and sorts the chunks by relevance, so that the most relevant chunk lies closest to the query. QGPM instead keeps every token in place and needs no external ranker. The tokenizer of the reranker is not fully compatible with those of Ministral-8B and Granite-3.3-8B, so we run this comparison on Qwen2.5-7B only.

Table~\ref{tab:ablation} reports the averages, and Figure~\ref{fig:edit-length} shows each context length. On NoLiMa, both variants outperform Vanilla and STRING, with 41.1 points for Top-K and 39.9 points for Reorder. Query relevance thus helps even without positional modulation. QGPM reaches 46.1 points and stays above both variants at every length, by 0.4 to 10.2 points. On NIAH-MV, QGPM also leads with 93.4 points, against 92.2 for Reorder and 91.1 for Top-K. The margin over Top-K reaches 5.2 points at 64K, likely because the five kept chunks do not always cover every value of the key. Top-K loses any evidence that the ranker misses, and Reorder breaks the original order of the document. QGPM avoids both risks because every token stays in the input. On QA-1, the three methods perform on par, and Top-K is 0.2 points above QGPM. Unlike NoLiMa, QA-1 does not minimize the lexical overlap between question and evidence, so the ranker likely keeps the evidence and deletion costs little.

\subsection{Quality of Head Relevance (RQ3)}
\label{sec:exp-oracle}

Oracle replaces the relevance read from attention heads with evidence positions marked by hand from the benchmark labels. All other steps stay unchanged, including the two maps, the protected prefix, and the frequency gate. Any gap between QGPM and Oracle therefore comes from the relevance signal alone. Marking the evidence by hand is costly, so we run Oracle on Qwen2.5-7B only.

On all three tasks of Qwen2.5, QGPM comes within 1.2 points of Oracle (Table~\ref{tab:ablation}). The relevance read from attention heads is thus close to annotated evidence. QGPM even exceeds Oracle at some lengths, such as 8K on NoLiMa and 64K on NIAH-MV (Figure~\ref{fig:edit-length}). Oracle is therefore a reference rather than a strict upper bound.

On NoLiMa and QA-1, the remaining gap concentrates at the longest length. On NoLiMa, Oracle improves on Vanilla by 17.6 to 21.2 points at every length. Moving correctly identified evidence closer to the query thus helps even at 32K. QGPM trails Oracle by at most 2.0 points up to 16K, but the gap grows to 5.7 points at 32K. At this length, the relevance signal rather than the positional modulation limits QGPM. The smaller NoLiMa margin at 32K in Section~\ref{sec:exp-main} likely has the same cause. On QA-1, Oracle gains only 2.0 points over Vanilla on average, so position is not the main bottleneck for this task. The exception is again the longest length, where Oracle gains 8.0 points at 64K and QGPM gains 3.2 points.

\section{Conclusion}
\label{sec:conclusion}

We presented QGPM, a training-free method that reshapes RoPE positions using query relevance read from the model's own attention heads. QGPM compresses low-relevance text only on fast-rotating planes and guides the lower and upper layers with a broad and a sharp relevance map. Every token stays in the input, and no model weight changes. Across four backbones, QGPM achieves the best average accuracy, with the largest gains on retrieval tasks. On question answering, where position matters less, QGPM matches the backbone. Modulating positions also outperforms input editing on retrieval, and the head relevance comes close to annotated evidence. The remaining gap appears at the longest contexts, so better relevance estimation for very long inputs is a promising next step.

\section*{Limitations}

\paragraph{Access to model internals.}
QGPM reads the query and key vectors of selected heads and replaces the RoPE tables of the backbone. The method therefore applies to open-weight models but not to models served only through an API. Most open long-context models use RoPE, so this requirement still covers a broad range of backbones.
 
\paragraph{Extra prefill.}
QGPM adds one plain prefill that stops at the deepest selected layer, which increases the time to the first token. The decoding cost and the size of the KV cache stay unchanged. The context keys in this first pass do not depend on the question. Several questions over the same document could therefore share a cached prefix. A shallower head set would also shorten the first pass.

\section*{Ethics Statement}

This work proposes a training-free inference method that changes how a long-context language model encodes positions. QGPM introduces no new model, training data, or alignment procedure, so the safety and factuality of each backbone remain those of the released model. All experiments use public benchmarks, NoLiMa and RULER, which consist of synthetic or existing question answering instances without personal information. The backbones and the reranker used by the input editing baselines are publicly released models. QGPM requires no fine-tuning, so the computational cost of this work is modest. We are not aware of risks specific to QGPM beyond those of the underlying models.

\bibliography{custom}

\appendix

\section{Head Selection}
\label{app:heads}

QGPM reads relevance from a fixed set $\mathcal{H}$ of attention heads, selected once for each backbone. Every (layer, head) pair of the backbone is a candidate. The calibration set is a fixed subset of NoLiMa samples whose evidence spans are known. The calibration samples are disjoint from the evaluation samples.

For each calibration sample and candidate head, we compute the sentence signal $\ell_h(S_i)$ of Equation~\ref{eq:head-signal} with the same query rows as at inference. The context sentences are then ranked by this signal. The quality of a head is the mean reciprocal rank of the evidence sentence over the calibration samples, with ties counted at the expected rank. The $|\mathcal{H}|$ heads with the highest quality form the head set. Calibration only selects heads and updates no weight.

Three details keep the signal faithful to the attention of the model. Under grouped-query attention, head $h$ reads the key head of the same group. On Qwen3, the vectors are taken after the query and key normalization of the model. Tokens that contain a newline character are excluded from the signal, because the line break around a NoLiMa needle is a formatting artifact that some shallow heads attend to.

The selected heads lie mostly in the middle and upper layers of each backbone, which motivates the layer schedule in Section~\ref{sec:method-remapping}.

\section{Implementation Details}
\label{app:implementation}

\paragraph{Backbones.}
Table~\ref{tab:backbones} lists the configuration of the four backbones. All four use a head dimension of $d = 128$ and unscaled RoPE, so the boundary plane $m^*$ follows Equation~\ref{eq:wrap-index} directly.

\begin{table}[t]
\centering
\small
\setlength{\tabcolsep}{3pt}
\begin{tabular*}{\columnwidth}{@{\extracolsep{\fill}} l cccc @{}}
\toprule
Backbone & Layers & Heads & $\theta$ & Window \\
\midrule
Qwen2.5-7B & 28 & 28 / 4 & $10^{7}$ & 1,010,000 \\
Qwen3-4B & 36 & 32 / 8 & $5 \times 10^{6}$ & 262,144 \\
Ministral-8B & 36 & 32 / 8 & $10^{8}$ & 32,768 \\
Granite-3.3-8B & 40 & 32 / 8 & $10^{7}$ & 131,072 \\
\bottomrule
\end{tabular*}
\caption{Backbone configurations from the HuggingFace releases. Heads lists query heads and key-value heads. Window is the configured maximum number of positions.}
\label{tab:backbones}
\end{table}

\paragraph{Prompts and query rows.}
Each prompt contains a system prompt, a task instruction, the context, the question, and an answer prefix. Only the context is scored, and every other token receives a score of 1. On NoLiMa, the context is a book excerpt with one inserted needle. On RULER, the context is the haystack, and the question is the text from the last \texttt{Question:} onward. The query rows $\mathcal{Q}$ are the question tokens together with the last few tokens of the prompt. All methods read exactly the same prompts and needle placements.

\paragraph{Sentence segmentation.}
The context is split into sentences after ``.'', ``!'', and ``?'' followed by whitespace, with common abbreviations protected. A token belongs to the sentence that contains the first character of the token, and the whitespace between two sentences belongs to the next sentence. Tokens are aligned to sentences through the character offsets of the backbone tokenizer.

\paragraph{Position tables.}
The remapped positions and the cosine and sine tables are computed in FP32 and then cast to the model precision. QGPM builds the tables once per prefill for each map. A forward pre-hook passes the tables of the assigned map to the attention of each decoder layer, so the KV cache of each layer stays in the coordinates of that map. With every score equal to 1, the patched model reproduces the unmodified backbone, which we verify for each backbone.

\paragraph{Decoding and hardware.}
All methods use greedy decoding with the same limit on new tokens for each dataset. The experiments run with PyTorch and HuggingFace Transformers on NVIDIA A100 GPUs.

\paragraph{Metrics.}
Each response is scored as correct or incorrect, and accuracy is the percentage of correct responses. On NoLiMa, a response is correct if the response contains the gold answer. On RULER QA, a response must contain the gold answer, and a question with several required answers needs every answer. NIAH-MK requires the value of the target key, and NIAH-MV requires every value of the key. Each RULER task has 500 samples at each length.

\paragraph{Baselines.}
All baselines use the default settings recommended in the original papers.

\paragraph{Input editing.}
Top-K and Reorder split the context into sentence-aligned chunks and score each chunk against the question with Qwen3-Reranker-4B. Top-K keeps the five highest-scoring chunks and drops the rest. Reorder keeps all chunks and sorts the chunks by score, so that the most relevant chunk ends right before the question. Both variants run with standard RoPE. The tokenizer of the reranker is not fully compatible with those of Ministral-8B and Granite-3.3-8B, so both variants run on Qwen2.5-7B only.

\begin{table*}[t]
\centering
\small
\setlength{\tabcolsep}{3pt}
\begin{tabular*}{\textwidth}{@{\extracolsep{\fill}} ll cccccc @{\hspace{10pt}} cccccc @{}}
\toprule
 & & \multicolumn{6}{c}{Qwen2.5-7B} & \multicolumn{6}{c}{Qwen3-4B} \\
\cmidrule(lr){3-8} \cmidrule(l){9-14}
Task & Method & 4K & 8K & 16K & 32K & 64K & Avg. & 4K & 8K & 16K & 32K & 64K & Avg. \\
\midrule
\multirow{6}{*}{NoLiMa} & Vanilla & \phantom{0}36.7 & \phantom{0}32.9 & \phantom{0}27.2 & \phantom{0}17.2 & -- & \phantom{0}28.5 & \phantom{0}30.6 & \phantom{0}19.4 & \phantom{0}10.3 & \phantom{00}4.0 & -- & \phantom{0}16.1 \\
 & LLMLingua-2 & \phantom{0}29.8 & \phantom{0}18.5 & \phantom{0}17.4 & \phantom{0}13.2 & -- & \phantom{0}19.7 & \phantom{0}23.2 & \phantom{0}10.7 & \phantom{00}5.0 & \phantom{00}2.8 & -- & \phantom{0}10.4 \\
 & Ms-PoE & \phantom{0}41.2 & \phantom{0}36.6 & \phantom{0}32.2 & \phantom{0}21.7 & -- & \phantom{0}32.9 & \phantom{0}32.2 & \phantom{0}19.9 & \phantom{00}9.5 & \phantom{00}4.6 & -- & \phantom{0}16.6 \\
 & Self-Extend & \phantom{0}36.2 & \phantom{0}31.7 & \phantom{0}27.1 & \phantom{0}22.8 & -- & \phantom{0}29.5 & \phantom{0}30.7 & \phantom{0}19.9 & \phantom{0}13.9 & \phantom{00}8.0 & -- & \phantom{0}18.1 \\
 & STRING & \phantom{0}38.0 & \phantom{0}34.4 & \phantom{0}33.1 & \phantom{0}26.7 & -- & \phantom{0}33.1 & \phantom{0}35.6 & \phantom{0}29.5 & \phantom{0}18.7 & \phantom{00}7.9 & -- & \phantom{0}22.9 \\
 & QGPM & \textbf{\phantom{0}56.5} & \textbf{\phantom{0}54.9} & \textbf{\phantom{0}43.8} & \textbf{\phantom{0}29.1} & -- & \textbf{\phantom{0}46.1} & \textbf{\phantom{0}38.0} & \textbf{\phantom{0}33.2} & \textbf{\phantom{0}20.9} & \textbf{\phantom{00}8.3} & -- & \textbf{\phantom{0}25.1} \\
\midrule
\multirow{6}{*}{QA-1} & Vanilla & \phantom{0}86.4 & \textbf{\phantom{0}82.4} & \phantom{0}80.8 & \phantom{0}77.0 & \phantom{0}71.4 & \phantom{0}79.6 & \textbf{\phantom{0}85.2} & \phantom{0}80.0 & \phantom{0}77.0 & \phantom{0}75.4 & \phantom{0}69.8 & \phantom{0}77.5 \\
 & LLMLingua-2 & \phantom{0}56.2 & \phantom{0}52.8 & \phantom{0}50.2 & \phantom{0}49.4 & \phantom{0}46.6 & \phantom{0}51.0 & \phantom{0}53.2 & \phantom{0}49.8 & \phantom{0}50.2 & \phantom{0}48.4 & \phantom{0}43.8 & \phantom{0}49.1 \\
 & Ms-PoE & \phantom{0}83.0 & \phantom{0}79.0 & \phantom{0}77.4 & \phantom{0}75.8 & \phantom{0}71.0 & \phantom{0}77.2 & \phantom{0}84.2 & \phantom{0}77.8 & \phantom{0}76.0 & \phantom{0}74.6 & \phantom{0}64.4 & \phantom{0}75.4 \\
 & Self-Extend & \phantom{0}86.4 & \phantom{0}79.8 & \phantom{0}79.6 & \phantom{0}77.4 & \phantom{0}73.2 & \phantom{0}79.3 & \textbf{\phantom{0}85.2} & \phantom{0}79.8 & \phantom{0}75.6 & \phantom{0}75.6 & \phantom{0}66.2 & \phantom{0}76.5 \\
 & STRING & \phantom{0}85.0 & \phantom{0}82.0 & \phantom{0}77.2 & \textbf{\phantom{0}78.2} & \phantom{0}74.4 & \phantom{0}79.4 & \phantom{0}84.2 & \textbf{\phantom{0}80.4} & \phantom{0}75.6 & \phantom{0}77.4 & \textbf{\phantom{0}76.6} & \phantom{0}78.8 \\
 & QGPM & \textbf{\phantom{0}87.4} & \phantom{0}80.8 & \textbf{\phantom{0}81.6} & \phantom{0}77.4 & \textbf{\phantom{0}74.6} & \textbf{\phantom{0}80.4} & \phantom{0}84.6 & \textbf{\phantom{0}80.4} & \textbf{\phantom{0}81.8} & \textbf{\phantom{0}78.4} & \phantom{0}75.4 & \textbf{\phantom{0}80.1} \\
\midrule
\multirow{6}{*}{QA-2} & Vanilla & \textbf{\phantom{0}64.0} & \phantom{0}58.8 & \phantom{0}56.8 & \phantom{0}55.6 & \phantom{0}50.0 & \phantom{0}57.0 & \textbf{\phantom{0}62.0} & \phantom{0}60.2 & \phantom{0}57.0 & \phantom{0}52.6 & \phantom{0}51.4 & \phantom{0}56.6 \\
 & LLMLingua-2 & \phantom{0}41.0 & \phantom{0}38.8 & \phantom{0}41.6 & \phantom{0}40.8 & \phantom{0}37.6 & \phantom{0}40.0 & \phantom{0}37.0 & \phantom{0}38.0 & \phantom{0}37.2 & \phantom{0}35.2 & \phantom{0}35.6 & \phantom{0}36.6 \\
 & Ms-PoE & \phantom{0}58.4 & \phantom{0}54.8 & \phantom{0}51.4 & \phantom{0}51.8 & \phantom{0}43.8 & \phantom{0}52.0 & \phantom{0}58.0 & \phantom{0}61.0 & \phantom{0}55.0 & \phantom{0}52.6 & \phantom{0}50.0 & \phantom{0}55.3 \\
 & Self-Extend & \textbf{\phantom{0}64.0} & \phantom{0}59.8 & \phantom{0}58.4 & \phantom{0}54.2 & \phantom{0}51.2 & \phantom{0}57.5 & \textbf{\phantom{0}62.0} & \phantom{0}59.4 & \phantom{0}59.4 & \phantom{0}53.6 & \phantom{0}53.0 & \phantom{0}57.5 \\
 & STRING & \phantom{0}61.0 & \phantom{0}60.4 & \textbf{\phantom{0}59.0} & \phantom{0}56.2 & \phantom{0}52.2 & \phantom{0}57.8 & \phantom{0}60.0 & \textbf{\phantom{0}63.4} & \textbf{\phantom{0}62.0} & \textbf{\phantom{0}57.2} & \textbf{\phantom{0}57.6} & \textbf{\phantom{0}60.0} \\
 & QGPM & \phantom{0}61.4 & \textbf{\phantom{0}60.8} & \textbf{\phantom{0}59.0} & \textbf{\phantom{0}56.6} & \textbf{\phantom{0}52.4} & \textbf{\phantom{0}58.0} & \phantom{0}60.6 & \phantom{0}59.2 & \phantom{0}56.6 & \phantom{0}52.6 & \phantom{0}52.2 & \phantom{0}56.2 \\
\midrule
\multirow{6}{*}{NIAH-MK} & Vanilla & \textbf{100.0} & \phantom{0}97.2 & \phantom{0}92.8 & \phantom{0}90.2 & \phantom{0}91.4 & \phantom{0}94.3 & \textbf{\phantom{0}99.8} & \phantom{0}99.8 & \textbf{100.0} & \phantom{0}99.4 & \phantom{0}99.2 & \phantom{0}99.6 \\
 & LLMLingua-2 & \phantom{00}1.8 & \phantom{00}2.2 & \phantom{00}2.4 & \phantom{00}2.4 & \phantom{00}2.2 & \phantom{00}2.2 & \phantom{00}2.2 & \phantom{00}2.0 & \phantom{00}2.0 & \phantom{00}1.6 & \phantom{00}2.0 & \phantom{00}2.0 \\
 & Ms-PoE & \phantom{0}50.8 & \phantom{0}51.6 & \phantom{0}39.0 & \phantom{0}27.4 & \phantom{0}15.2 & \phantom{0}36.8 & \phantom{0}95.0 & \phantom{0}98.4 & \phantom{0}94.8 & \phantom{0}88.0 & \phantom{0}65.0 & \phantom{0}88.2 \\
 & Self-Extend & \textbf{100.0} & \phantom{0}96.0 & \phantom{0}78.6 & \phantom{0}61.2 & \phantom{0}58.0 & \phantom{0}78.8 & \textbf{\phantom{0}99.8} & \textbf{100.0} & \phantom{0}99.8 & \textbf{\phantom{0}99.8} & \textbf{\phantom{0}99.8} & \textbf{\phantom{0}99.8} \\
 & STRING & \phantom{0}98.4 & \phantom{0}96.2 & \phantom{0}95.0 & \phantom{0}84.8 & \phantom{0}87.2 & \phantom{0}92.3 & \phantom{0}99.4 & \phantom{0}99.6 & \phantom{0}99.2 & \phantom{0}99.6 & \phantom{0}98.6 & \phantom{0}99.3 \\
 & QGPM & \textbf{100.0} & \textbf{\phantom{0}99.8} & \textbf{\phantom{0}98.6} & \textbf{\phantom{0}99.2} & \textbf{\phantom{0}99.2} & \textbf{\phantom{0}99.4} & \textbf{\phantom{0}99.8} & \phantom{0}99.8 & \textbf{100.0} & \phantom{0}99.6 & \phantom{0}98.8 & \phantom{0}99.6 \\
\midrule
\multirow{6}{*}{NIAH-MV} & Vanilla & \phantom{0}85.0 & \phantom{0}82.4 & \phantom{0}82.2 & \phantom{0}86.8 & \phantom{0}81.2 & \phantom{0}83.5 & \phantom{0}93.8 & \phantom{0}95.4 & \phantom{0}92.2 & \textbf{\phantom{0}99.2} & \textbf{\phantom{0}99.6} & \phantom{0}96.0 \\
 & LLMLingua-2 & \phantom{0}69.8 & \phantom{0}65.8 & \phantom{0}66.8 & \phantom{0}63.6 & \phantom{0}60.6 & \phantom{0}65.3 & \phantom{0}76.6 & \phantom{0}76.4 & \phantom{0}79.0 & \phantom{0}74.8 & \phantom{0}72.6 & \phantom{0}75.9 \\
 & Ms-PoE & \phantom{0}79.2 & \phantom{0}56.2 & \phantom{0}62.8 & \phantom{0}71.0 & \phantom{0}73.4 & \phantom{0}68.5 & \phantom{0}79.2 & \phantom{0}85.2 & \phantom{0}81.4 & \phantom{0}86.0 & \phantom{0}91.8 & \phantom{0}84.7 \\
 & Self-Extend & \phantom{0}85.2 & \phantom{0}82.4 & \phantom{0}84.0 & \phantom{0}77.2 & \phantom{0}84.2 & \phantom{0}82.6 & \phantom{0}94.0 & \phantom{0}91.6 & \phantom{0}88.8 & \phantom{0}93.2 & \phantom{0}93.6 & \phantom{0}92.2 \\
 & STRING & \phantom{0}79.8 & \phantom{0}77.2 & \phantom{0}76.4 & \phantom{0}75.2 & \phantom{0}74.2 & \phantom{0}76.6 & \phantom{0}94.2 & \phantom{0}91.6 & \phantom{0}89.6 & \phantom{0}92.8 & \phantom{0}96.6 & \phantom{0}93.0 \\
 & QGPM & \textbf{\phantom{0}93.4} & \textbf{\phantom{0}90.2} & \textbf{\phantom{0}92.0} & \textbf{\phantom{0}95.8} & \textbf{\phantom{0}95.4} & \textbf{\phantom{0}93.4} & \textbf{\phantom{0}96.4} & \textbf{\phantom{0}96.2} & \textbf{\phantom{0}93.4} & \phantom{0}98.8 & \phantom{0}99.4 & \textbf{\phantom{0}96.8} \\
\bottomrule
\end{tabular*}
\caption{Accuracy (\%) at each context length on Qwen2.5-7B and Qwen3-4B. NoLiMa is evaluated up to 32K, and Avg.\ is the mean over the evaluated lengths. Bold marks the best result for each backbone, task, and length.}
\label{tab:full-qwen}
\end{table*}

\begin{table*}[t]
\centering
\small
\setlength{\tabcolsep}{3pt}
\begin{tabular*}{\textwidth}{@{\extracolsep{\fill}} ll cccccc @{\hspace{10pt}} cccccc @{}}
\toprule
 & & \multicolumn{6}{c}{Ministral-8B} & \multicolumn{6}{c}{Granite-3.3-8B} \\
\cmidrule(lr){3-8} \cmidrule(l){9-14}
Task & Method & 4K & 8K & 16K & 32K & 64K$^{\dagger}$ & Avg. & 4K & 8K & 16K & 32K & 64K & Avg. \\
\midrule
\multirow{6}{*}{NoLiMa} & Vanilla & \phantom{0}25.8 & \phantom{0}13.0 & \phantom{00}4.5 & \phantom{00}2.6 & -- & \phantom{0}11.5 & \phantom{0}31.8 & \phantom{0}21.7 & \phantom{0}12.9 & \phantom{00}7.9 & -- & \phantom{0}18.6 \\
 & LLMLingua-2 & \phantom{0}15.9 & \phantom{00}6.1 & \phantom{00}1.9 & \phantom{00}0.8 & -- & \phantom{00}6.2 & \phantom{0}22.8 & \phantom{0}16.9 & \phantom{0}10.3 & \phantom{00}6.5 & -- & \phantom{0}14.1 \\
 & Ms-PoE & \phantom{0}19.8 & \phantom{00}9.6 & \phantom{00}3.6 & \phantom{00}2.9 & -- & \phantom{00}9.0 & \phantom{0}31.6 & \phantom{0}19.8 & \phantom{0}14.4 & \phantom{00}8.3 & -- & \phantom{0}18.5 \\
 & Self-Extend & \phantom{0}25.8 & \phantom{0}11.5 & \phantom{00}4.4 & \phantom{00}2.5 & -- & \phantom{0}11.1 & \phantom{0}29.0 & \phantom{0}18.7 & \phantom{0}12.0 & \phantom{00}6.5 & -- & \phantom{0}16.6 \\
 & STRING & \phantom{0}32.9 & \phantom{0}19.8 & \phantom{0}10.6 & \phantom{00}5.1 & -- & \phantom{0}17.1 & \phantom{0}40.3 & \phantom{0}30.0 & \phantom{0}20.8 & \phantom{0}12.4 & -- & \phantom{0}25.9 \\
 & QGPM & \textbf{\phantom{0}46.3} & \textbf{\phantom{0}32.6} & \textbf{\phantom{0}12.3} & \textbf{\phantom{00}5.4} & -- & \textbf{\phantom{0}24.1} & \textbf{\phantom{0}52.5} & \textbf{\phantom{0}39.8} & \textbf{\phantom{0}28.9} & \textbf{\phantom{0}21.3} & -- & \textbf{\phantom{0}35.6} \\
\midrule
\multirow{6}{*}{QA-1} & Vanilla & \phantom{0}80.0 & \textbf{\phantom{0}77.0} & \phantom{0}75.0 & \phantom{0}72.0 & \phantom{0}63.6 & \phantom{0}73.5 & \phantom{0}83.6 & \textbf{\phantom{0}81.0} & \phantom{0}78.6 & \phantom{0}79.0 & \phantom{0}70.6 & \phantom{0}78.6 \\
 & LLMLingua-2 & \phantom{0}50.8 & \phantom{0}51.0 & \phantom{0}46.8 & \phantom{0}46.8 & \phantom{0}45.4 & \phantom{0}48.2 & \phantom{0}53.8 & \phantom{0}48.8 & \phantom{0}51.2 & \phantom{0}49.8 & \phantom{0}47.0 & \phantom{0}50.1 \\
 & Ms-PoE & \phantom{0}77.2 & \phantom{0}73.0 & \phantom{0}69.2 & \phantom{0}66.2 & \phantom{0}56.8 & \phantom{0}68.5 & \phantom{0}82.0 & \phantom{0}77.4 & \phantom{0}74.8 & \phantom{0}74.4 & \phantom{0}61.8 & \phantom{0}74.1 \\
 & Self-Extend & \phantom{0}80.0 & \phantom{0}76.2 & \phantom{0}74.0 & \phantom{0}71.8 & \phantom{0}62.0 & \phantom{0}72.8 & \phantom{0}85.0 & \phantom{0}78.4 & \phantom{0}77.0 & \phantom{0}76.0 & \phantom{0}70.0 & \phantom{0}77.3 \\
 & STRING & \phantom{0}79.6 & \phantom{0}76.0 & \phantom{0}75.6 & \textbf{\phantom{0}72.2} & \phantom{0}63.0 & \phantom{0}73.3 & \phantom{0}82.0 & \phantom{0}80.0 & \phantom{0}81.0 & \textbf{\phantom{0}80.2} & \phantom{0}70.8 & \textbf{\phantom{0}78.8} \\
 & QGPM & \textbf{\phantom{0}80.4} & \phantom{0}75.4 & \textbf{\phantom{0}76.8} & \phantom{0}71.8 & \textbf{\phantom{0}63.8} & \textbf{\phantom{0}73.6} & \textbf{\phantom{0}87.2} & \phantom{0}77.8 & \textbf{\phantom{0}81.4} & \phantom{0}76.8 & \textbf{\phantom{0}71.0} & \textbf{\phantom{0}78.8} \\
\midrule
\multirow{6}{*}{QA-2} & Vanilla & \textbf{\phantom{0}66.0} & \phantom{0}62.8 & \phantom{0}61.4 & \textbf{\phantom{0}59.4} & \phantom{0}45.6 & \phantom{0}59.0 & \phantom{0}66.4 & \phantom{0}64.2 & \phantom{0}63.4 & \phantom{0}61.8 & \textbf{\phantom{0}57.6} & \phantom{0}62.7 \\
 & LLMLingua-2 & \phantom{0}41.2 & \phantom{0}42.6 & \phantom{0}43.6 & \phantom{0}37.6 & \phantom{0}38.8 & \phantom{0}40.8 & \phantom{0}45.0 & \phantom{0}42.6 & \phantom{0}44.8 & \phantom{0}41.8 & \phantom{0}41.4 & \phantom{0}43.1 \\
 & Ms-PoE & \phantom{0}58.6 & \phantom{0}60.6 & \phantom{0}55.2 & \phantom{0}50.0 & \phantom{0}40.6 & \phantom{0}53.0 & \phantom{0}65.0 & \phantom{0}62.8 & \phantom{0}60.2 & \phantom{0}62.0 & \phantom{0}51.0 & \phantom{0}60.2 \\
 & Self-Extend & \phantom{0}63.8 & \textbf{\phantom{0}63.6} & \phantom{0}61.4 & \phantom{0}59.2 & \phantom{0}46.2 & \phantom{0}58.8 & \phantom{0}67.2 & \phantom{0}65.4 & \phantom{0}60.8 & \phantom{0}60.4 & \phantom{0}55.6 & \phantom{0}61.9 \\
 & STRING & \phantom{0}65.0 & \phantom{0}62.8 & \textbf{\phantom{0}62.2} & \phantom{0}59.2 & \phantom{0}46.4 & \textbf{\phantom{0}59.1} & \phantom{0}67.4 & \phantom{0}66.2 & \textbf{\phantom{0}64.4} & \phantom{0}61.2 & \phantom{0}55.4 & \phantom{0}62.9 \\
 & QGPM & \phantom{0}64.4 & \phantom{0}62.4 & \phantom{0}60.8 & \textbf{\phantom{0}59.4} & \textbf{\phantom{0}48.2} & \phantom{0}59.0 & \textbf{\phantom{0}67.6} & \textbf{\phantom{0}67.2} & \phantom{0}63.6 & \textbf{\phantom{0}62.6} & \phantom{0}54.6 & \textbf{\phantom{0}63.1} \\
\midrule
\multirow{6}{*}{NIAH-MK} & Vanilla & \textbf{100.0} & \textbf{100.0} & \phantom{0}99.6 & \phantom{0}98.6 & \phantom{0}49.0 & \phantom{0}89.4 & \phantom{0}95.4 & \phantom{0}89.4 & \phantom{0}79.6 & \phantom{0}51.2 & \phantom{0}14.2 & \phantom{0}66.0 \\
 & LLMLingua-2 & \phantom{00}3.2 & \phantom{00}3.4 & \phantom{00}3.6 & \phantom{00}3.6 & \phantom{00}4.2 & \phantom{00}3.6 & \phantom{00}0.2 & \phantom{00}0.4 & \phantom{00}0.0 & \phantom{00}0.4 & \phantom{00}0.0 & \phantom{00}0.2 \\
 & Ms-PoE & \phantom{0}50.2 & \phantom{0}50.4 & \phantom{0}30.6 & \phantom{0}20.4 & \phantom{0}10.6 & \phantom{0}32.4 & \phantom{0}89.2 & \phantom{0}85.4 & \phantom{0}74.8 & \phantom{0}42.0 & \phantom{0}16.2 & \phantom{0}61.5 \\
 & Self-Extend & \textbf{100.0} & \phantom{0}99.8 & \phantom{0}99.6 & \phantom{0}98.2 & \phantom{0}75.2 & \phantom{0}94.6 & \phantom{0}95.2 & \phantom{0}87.4 & \phantom{0}61.4 & \phantom{0}29.0 & \phantom{0}11.2 & \phantom{0}56.8 \\
 & STRING & \textbf{100.0} & \phantom{0}99.8 & \phantom{0}99.8 & \phantom{0}98.2 & \phantom{0}75.8 & \phantom{0}94.7 & \phantom{0}92.0 & \phantom{0}89.8 & \phantom{0}83.2 & \phantom{0}63.6 & \phantom{0}39.0 & \phantom{0}73.5 \\
 & QGPM & \textbf{100.0} & \textbf{100.0} & \textbf{100.0} & \textbf{\phantom{0}99.6} & \textbf{\phantom{0}76.4} & \textbf{\phantom{0}95.2} & \textbf{\phantom{0}98.2} & \textbf{\phantom{0}98.2} & \textbf{\phantom{0}96.6} & \textbf{\phantom{0}77.0} & \textbf{\phantom{0}55.4} & \textbf{\phantom{0}85.1} \\
\midrule
\multirow{6}{*}{NIAH-MV} & Vanilla & \phantom{0}98.4 & \phantom{0}98.4 & \phantom{0}96.6 & \phantom{0}93.6 & \phantom{0}69.8 & \phantom{0}91.4 & \phantom{0}82.6 & \phantom{0}74.2 & \phantom{0}68.8 & \phantom{0}68.8 & \phantom{0}61.0 & \phantom{0}71.1 \\
 & LLMLingua-2 & \phantom{0}83.2 & \phantom{0}82.2 & \phantom{0}82.2 & \phantom{0}75.2 & \phantom{0}67.4 & \phantom{0}78.0 & \phantom{0}64.4 & \phantom{0}56.2 & \phantom{0}52.4 & \phantom{0}46.6 & \phantom{0}22.6 & \phantom{0}48.4 \\
 & Ms-PoE & \phantom{0}81.6 & \phantom{0}91.6 & \phantom{0}87.8 & \phantom{0}80.6 & \phantom{0}64.2 & \phantom{0}81.2 & \phantom{0}20.4 & \phantom{00}9.8 & \phantom{0}23.6 & \phantom{0}13.0 & \phantom{00}3.8 & \phantom{0}14.1 \\
 & Self-Extend & \phantom{0}98.4 & \phantom{0}96.8 & \phantom{0}96.6 & \phantom{0}94.6 & \phantom{0}77.8 & \phantom{0}92.8 & \phantom{0}83.0 & \phantom{0}78.2 & \phantom{0}75.0 & \phantom{0}64.8 & \phantom{0}58.0 & \phantom{0}71.8 \\
 & STRING & \phantom{0}95.8 & \phantom{0}92.6 & \phantom{0}96.4 & \phantom{0}94.0 & \phantom{0}75.2 & \phantom{0}90.8 & \phantom{0}84.2 & \phantom{0}76.2 & \phantom{0}80.6 & \phantom{0}68.8 & \phantom{0}69.4 & \phantom{0}75.8 \\
 & QGPM & \textbf{\phantom{0}98.8} & \textbf{\phantom{0}98.6} & \textbf{\phantom{0}96.8} & \textbf{\phantom{0}94.8} & \textbf{\phantom{0}78.2} & \textbf{\phantom{0}93.4} & \textbf{\phantom{0}84.4} & \textbf{\phantom{0}79.0} & \textbf{\phantom{0}81.2} & \textbf{\phantom{0}80.0} & \textbf{\phantom{0}70.0} & \textbf{\phantom{0}78.9} \\
\bottomrule
\end{tabular*}
\caption{Accuracy (\%) at each context length on Ministral-8B and Granite-3.3-8B. NoLiMa is evaluated up to 32K, and Avg.\ is the mean over the evaluated lengths. Bold marks the best result for each backbone, task, and length. $^{\dagger}$Exceeds the configured window of Ministral-8B.}
\label{tab:full-other}
\end{table*}

\begin{table}[t]
\centering
\small
\setlength{\tabcolsep}{2.5pt}
\begin{tabular*}{\columnwidth}{@{\extracolsep{\fill}} ll cccccc @{}}
\toprule
Task & Method & 4K & 8K & 16K & 32K & 64K & Avg. \\
\midrule
\multirow{6}{*}{NoLiMa} & Vanilla & 36.7 & 32.9 & 27.2 & 17.2 & -- & 28.5 \\
 & STRING & 38.0 & 34.4 & 33.1 & 26.7 & -- & 33.1 \\
 & Top-K & 47.9 & 50.7 & 37.2 & 28.7 & -- & 41.1 \\
 & Reorder & 50.4 & 44.7 & 38.2 & 26.4 & -- & 39.9 \\
 & QGPM & \textbf{56.5} & \textbf{54.9} & \textbf{43.8} & \textbf{29.1} & -- & \textbf{46.1} \\
 & \textcolor{gray}{Oracle} & \textcolor{gray}{57.9} & \textcolor{gray}{50.8} & \textcolor{gray}{45.8} & \textcolor{gray}{34.8} & \textcolor{gray}{--} & \textcolor{gray}{47.3} \\
\midrule
\multirow{6}{*}{QA-1} & Vanilla & 86.4 & \textbf{82.4} & 80.8 & 77.0 & 71.4 & 79.6 \\
 & STRING & 85.0 & 82.0 & 77.2 & 78.2 & 74.4 & 79.4 \\
 & Top-K & 86.2 & 82.2 & 81.0 & \textbf{79.6} & 74.0 & \textbf{80.6} \\
 & Reorder & 84.8 & 81.2 & 81.2 & 79.2 & 74.0 & 80.1 \\
 & QGPM & \textbf{87.4} & 80.8 & \textbf{81.6} & 77.4 & \textbf{74.6} & 80.4 \\
 & \textcolor{gray}{Oracle} & \textcolor{gray}{85.8} & \textcolor{gray}{81.6} & \textcolor{gray}{83.0} & \textcolor{gray}{78.0} & \textcolor{gray}{79.4} & \textcolor{gray}{81.6} \\
\midrule
\multirow{6}{*}{NIAH-MV} & Vanilla & 85.0 & 82.4 & 82.2 & 86.8 & 81.2 & 83.5 \\
 & STRING & 79.8 & 77.2 & 76.4 & 75.2 & 74.2 & 76.6 \\
 & Top-K & 92.2 & \textbf{90.2} & 90.6 & 92.4 & 90.2 & 91.1 \\
 & Reorder & 92.8 & \textbf{90.2} & 91.2 & 93.2 & 93.8 & 92.2 \\
 & QGPM & \textbf{93.4} & \textbf{90.2} & \textbf{92.0} & \textbf{95.8} & \textbf{95.4} & \textbf{93.4} \\
 & \textcolor{gray}{Oracle} & \textcolor{gray}{93.6} & \textcolor{gray}{92.6} & \textcolor{gray}{92.8} & \textcolor{gray}{96.0} & \textcolor{gray}{94.6} & \textcolor{gray}{93.9} \\
\bottomrule
\end{tabular*}
\caption{Accuracy (\%) of QGPM, input editing, and Oracle at each context length on Qwen2.5-7B. Oracle (gray) is shown for reference. Bold marks the best result other than Oracle.}
\label{tab:full-ablation}
\end{table}

\paragraph{Oracle.}
Oracle marks the evidence in each context by hand according to the benchmark labels, such as the needle in NoLiMa and the gold documents and inserted values in RULER. Oracle marks the gold spans at the token level, whereas the attention scorer of QGPM ranks whole sentences. The two maps, the protected prefix, the frequency gate, and the stride floor stay the same as in QGPM. Because these settings act on both, Oracle is a reference rather than a strict upper bound. Marking evidence by hand is costly, so we run Oracle on Qwen2.5-7B only.

\section{Hyperparameters}
\label{app:hyperparams}

QGPM sets the following hyperparameters for each backbone and dataset. Two settings are shared by all runs. The split layer is $\ell_s = \lfloor n_{\mathrm{layer}}/2 \rfloor$, and the score of the top sentence is 1. The remaining settings play the following roles.
\begin{itemize}
    \setlength{\itemsep}{0pt}
    \setlength{\parskip}{0pt}
    \item The broad map uses $(K_b, T_b)$ and the sharp map uses $(K_s, T_s)$, with $K_b > K_s$ and $T_b > T_s$. $K_s$ is at least the number of evidence spans that a task requires.
    \item The minimum score $s_{\mathrm{low}}$ and the stride floor $\epsilon$ set the strength of compression. A low-relevance token advances by $\max(s_{\mathrm{low}}, \epsilon)$ on a fast plane.
    \item The anchor length is $L_a = c \cdot L$, where $L$ is the context length and $c$ is set for each dataset. For the same $L_a$, a larger $\theta$ gives a smaller $m^*$, so fewer planes are compressed.
    \item The gate width $\tau_m$ is a small value in units of plane index.
\end{itemize}
On NIAH-MK and NIAH-MV, $K_s$ is chosen on development samples for each task and fixed before the evaluation samples are run.

\section{Computational Cost}
\label{app:cost}

QGPM changes the cost of inference in three places.
\begin{itemize}
    \setlength{\itemsep}{0pt}
    \setlength{\parskip}{0pt}
    \item The first pass is a plain prefill that stops right after the key projection of the deepest layer in $\mathcal{H}$. This pass costs at most one plain prefill and becomes cheaper with a shallower head set.
    \item The answering pass is a full prefill with the remapped tables. The attention kernels stay unchanged, so this pass costs the same as a plain prefill. Building the position tables takes $O(N d/2)$ time and memory for each map.
    \item Decoding keeps the cost per step and the KV cache size of standard inference. Each step adds one vector of $d/2$ angles for each map.
\end{itemize}

\section{Full Results}
\label{app:full-results}

Tables~\ref{tab:full-qwen} and~\ref{tab:full-other} report the accuracy of every method at each context length, and Table~\ref{tab:full-ablation} reports the comparison with input editing and Oracle at each length. The averages over lengths match Tables~\ref{tab:main} and~\ref{tab:ablation}.

\end{document}